\documentclass[11pt]{article}
\usepackage{geometry}
\usepackage{authblk}
\usepackage{graphicx}
\usepackage{coling2020}
\usepackage{framed}
\usepackage{times}
\usepackage{url}
\usepackage{latexsym}
\usepackage{microtype}
\usepackage{multirow}
\usepackage{makecell}
\usepackage{tabularx,booktabs}
\newcolumntype{Y}{>{\centering\arraybackslash}X}

\colingfinalcopy 

\title{UPB at SemEval-2020 Task 9: Identifying Sentiment in Code-Mixed Social Media Texts using Transformers and Multi-Task Learning}

\author{George-Eduard Zaharia, George-Alexandru Vlad, Dumitru-Clementin Cercel, \\ Traian Rebedea, Costin-Gabriel Chiru \\
   University Politehnica of Bucharest, Faculty of Automatic Control and Computers \\
  {\tt \{george.zaharia0806,george.vlad0108\}@stud.acs.pub.ro} \\
  \tt\{dumitru.cercel,traian.rebedea,costin.chiru\}@upb.ro}

\date{}
\begin{document}
\maketitle
\begin{abstract}
Sentiment analysis is a process widely used in opinion mining campaigns conducted today.
This phenomenon presents applications in a variety of fields, especially in collecting information related to the attitude or satisfaction of users concerning a particular subject. However, the task of managing such a process becomes noticeably more difficult when it is applied in cultures that tend to combine two languages in order to express ideas and thoughts. By interleaving words from two languages, the user can express with ease, but at the cost of making the text far less intelligible for those who are not familiar with this technique, but also for standard opinion mining algorithms.
In this paper, we describe the systems developed by our team for SemEval-2020 Task 9 that aims to cover two well-known code-mixed languages: Hindi-English and Spanish-English.
 We intend to solve this issue by introducing a solution that takes advantage of several neural network approaches, as well as pre-trained word embeddings.
Our approach (multlingual BERT) achieves promising performance on the Hindi-English task, with an average F1-score of 0.6850, registered on the competition leaderboard, ranking our team \(16^{th}\) out of 62 participants.
For the Spanish-English task, we obtained an  average F1-score of 0.7064 ranking our team \(17^{th}\) out of 29 participants by using another multilingual Transformer-based model, XLM-RoBERTa.
\end{abstract}

\section{Introduction}
\label{intro}

\blfootnote{This work is licensed under a Creative Commons Attribution 4.0 International Licence. Licence details: http://creativecommons.org/licenses/by/4.0/.
}

A series of countries from our world are multilingual, which implies that there are multiple languages spoken by their population. People tend to mix them at the phrase or sentence level in order to express ideas with ease, thus creating a phenomenon called code-mixing or code-switching. As it is expected, this embedding of a language into another one makes its appearance in the virtual space, as well. For example, Twitter users combine Hindi or Spanish phrases with English words, thus creating a bilingual phrase that can lead to understanding problems for non-natives.

However, the virtual space adds more layers of difficulty in identifying the sentiment of the author. Usually, social media users tend to adopt phonetic typing, which implies that the words will not take their original form, they will be adapted such that it will be faster to express the main idea. 
As a particular case, for the Hindi-English users, a new problem arises: Hindi and English use different alphabets, which determines the user to romanize the Hindi words such that both languages will use the same alphabet throughout the text. At the same time, social media users tend to express their sentiments by repeating certain vocals in words.  Furthermore, they use emojis, which will add an extra layer of complexity for analyzing the text.

This entire process creates new opportunities for research, given the importance of sentiment analysis in this area. The SemEval-2020 Task 9:  Sentiment Analysis for Code-Mixed Social Media Text~\cite{patwa2020sentimix} challenges the research community to solve the previously mentioned problem by introducing two subtasks, focusing on three of the world's most spoken languages\footnote{\url{https://www.visualcapitalist.com/100-most-spoken-languages/}}: Hindi and Spanish, alongside English. We proposed a series of neural models that intend to solve this issue, contributing under the usernames \textit{eduardgzaharia} and \textit{clementincercel}, respectively. Firstly, we experimented with Recurrent Neural Network (RNN)~\cite{Rumelhart1986LearningIR} solutions alongside word embeddings. After that, we performed the leap towards Transformer-based~\cite{vaswani2017attention} models that usually perform better and offer more insight for the combined language models. Furthermore, adding an auxiliary task for training a multi-task learning (MTL)~\cite{ruder2017overview} architecture can lead to even better results, as the models become able to learn new features from the input texts.

The paper is structured as follows. In section 2, we perform an analysis of existing solutions for several code-mixed tasks and sentiment analysis. In section 3, we detail the proposed approaches for code-mixed sentiment analysis. Section 4 details the performed experiments, including data and preprocessing, experimental setup, and a discussion of the results. Finally, we draw conclusions in section 5.

\section{Related Work}

\newcite{wang:2016} proposed a solution based on attention mechanisms for emotion prediction in code-switched texts. They introduced a new network architecture, namely Bilingual Attention Network consisting of two components. Firstly, a Long Short-term Memory (LSTM)~\cite{hochreiter:1997} network is used for obtaining the representation of the texts. Secondly, the system contains two types of attention, a monolingual one, which measures the importance of meaningful words from the monolingual contexts, and a bilingual attention that captures the priority of the most prominent groups of words, in order to create the bilingual attention vector.
The authors used Chinese-English code-mixed social media posts that are annotated with five classes for emotions: happiness, sadness, fear, anger, and surprise. They managed to improve the baseline scores by a margin of up to 7\% in accuracy.

MTL represents a distinct approach used for training neural networks for detecting the stance of the user in code-mixed texts~\cite{sushmita:2019}. The authors represented the main task as a classification with three stance classes (neutral, favor, against). At the same time, the auxiliary task is binary: either the author is neutral or has an opinion, positive or negative. 
Their approach obtains an accuracy of 63.2\% on the benchmark dataset proposed by \newcite{swami2018englishhindi}.

\newcite{trivedi:2018} introduced a multi-component neural network based on both word and character representations, with the purpose of performing Named Entity Recognition (NER) on code-switched data. The input words are defined as pre-trained fastText~\cite{bojanowski2016enriching} embeddings. A Convolutional Neural Network (CNN)~\cite{fukushima:neocognitronbc} has the role of learning character-level representations for words. 
More, another component is a Bidirectional LSTM (BiLSTM), used for learning the contextual relationships between the words in the same sentence. Their performance is over the considered baseline, surpassing it by a percentage of roughly 4\% in accuracy.

\newcite{mandal:2018} proposed an architecture suitable for NLP tasks aiming to tackle the language identification problem in code-mixed texts, called Multichannel Neural Network (MNN) that works alongside a BiLSTM-CRF. The MNN module is based on the Inception architecture \cite{szegedy2015rethinking}
and consists of four channels, one of them entering a LSTM, while the other three being connected to 1-D convolutional networks. The convolutional layers are used for capturing word representations at the n-gram level.
The BiLSTM-CRF module is directly connected to the previous one, with the purpose of learning the probabilities of switching language mid-sentence, based on the character embeddings. After training the architecture on two code-mixed datasets (Hindi-English and Bengali-English), their results surpassed the accuracy of the baseline model by 5\%.

For performing sentiment analysis, \newcite{vilares:2015} used an English-Spanish code-switching dataset containing 3,062 tweets, as well as two monolingual corpora from SemEval-2014~\cite{rosenthal2019semeval2014} and TASS-2014~\cite{Roman2014TASS2}.
At the same time, the authors used a multilingual corpus obtained by merging the previously mentioned monolingual sets. 
Their approach for identifying the sentiment implies extracting features by considering four atomic sets: words, lemmas, psychometric properties, and part-of-speech tags. By using this solution, the authors obtained excellent results: 98.12\% for the monolingual English dataset, 96.03\% for Spanish, as well as 98\% for the multilingual dataset.

\section{Proposed Solutions}
\subsection {Multilingual Word Embeddings}
Today, word embeddings represent a crucial aspect when performing natural language processing tasks. A word embedding represents an N-dimensional vector, which encodes the meaning of a certain word. Usually, the vector is created by analyzing the context in which the word is identified. The vector's values are adapted with every new encounter of a certain word, such that its meaning will be properly captured in every context. As a result, two similar words, having almost the same meaning (for example, "similar" and "related") will have the two representation vectors close to each other. Also, unrelated words will lead to distant vectors.

In order to speed up the training process, we have used pre-trained word embeddings. Thus, we were able to give the model an idea about the context of the words even before going through the dataset once. The embeddings were obtained using the 300-dimensional version of \textit{fastText}~\cite{bojanowski2016enriching}. However, the Hindi embedding vectors were associated with words written in the Hindi alphabet. Therefore, we needed to alter the embedding file by iterating over the words and romanizing each one of them, and then creating a new file with the romanized content alongside the vectors.

Furthermore, we encountered an additional problem: the word embeddings for English, Hindi, and Spanish, respectively, were trained separately, on different datasets. Therefore, the dimensions of the vectors from the embedding files were not aligned in the same vectorial space. We have used \textbf{MUSE} \cite{conneau2017word} to map one vectorial space into the other such that words with similar meaning from both languages are represented with extremely close vectors. MUSE is a tool which allows us, by providing two separate embedding files from two separate vectorial spaces, alongside a parallel corpora between the two languages, to map the vectors of one language into the space of the other. The MUSE repository\footnote{\url{https://github.com/facebookresearch/MUSE}} offers an English-Hindi, as well as an English-Spanish, parallel corpus. The English-Hindi corpus needed some tweaking, considering the fact that we were required to romanize the Hindi words once again. After running the tool, we managed to map the Hindi and Spanish embeddings into the space of the English embeddings. Therefore, words with the same meaning will have similar representations, being suitable to be used in code-mixed situations.

Similarly, we experimented with multilingual \textit{Transformer}-based embeddings from multilingual BERT (mBERT) \cite{pires:2019} and XLM-RoBERTa \cite{goyal:2020}. These are more powerful systems, considering the larger amounts of pre-training data, including Hindi, Spanish, and also English texts. The models allow a more precise representation of words in context because they use advanced encoding mechanisms based on self-attention \cite{vaswani2017attention}. Compared to the previous 300-dimensional embeddings, the Transformer models output, for the base variant, a 768-dimensional representation of a token, fact that adds even more precision.

\subsection{Neural Architectures}
\subsubsection{Bidirectional Long Short-Term Memory}
LSTM networks are RNNs used for preserving the sequentiality of data. We used a Bidirectional LSTM for learning the sentiment from the tweet. Being bidirectional, the tweet is fed once from the beginning to the end, and once more from the end to the beginning.

The BiLSTM takes vector input sequences and performs different operations which will lead to a series of annotated sequences, both in the forward and backward direction. For the final representation, the BiLSTM performs a concatenation of the forward and backward results. The annotations come in two dimensions, one for the timestep, the position of the word in the sequence, and the other for a certain neuron in the BiLSTM's hidden layer.

To avoid over-fitting, we use a \textit{dropout} layer which randomly sets the output of a percentage of neurons to be 0. The \textit{dropout} layer is followed by a 1D \textit{convolutional} layer, as well as a \textit{MaxPooling} layer, used for down-sampling the data. The output is a \textit{dense} layer with three neurons, one for each class.

\subsubsection{Gated Recurrent Units (GRUs)}
The vanishing gradient problem ~\cite{pascanu2012difficulty} in RNNs proves to be harmful to the overall performance of the model.  The GRUs~\cite{cho:2014} intend to solve this problem. They use an update gate and a reset gate that have the main purpose of deciding which information passes and which information is discarded. GRUs can also store old information found in the training set, information that is relevant for the overall performance. The update gate decides the quantity of already identified information that will be able to pass through the network, while the reset gate decides which information will be discarded.

\subsubsection{Capsule Networks}

While GRUs solve some of the problems of RNNs, Capsule Networks~\cite{e2018matrix} aim to solve problems of CNNs. Capsules learn to recognize entities and encode them into vectors. Because they are represented as vectors, the capsules require the existence of a function called squashing.
A Capsule Network\footnote{ \url{https://www.kaggle.com/fizzbuzz/beginner-s-guide-to-capsule-networks/notebook}} can contain N layers, each layer is dedicated to a more general aspect of a task. Deeper layers have a higher specificity or specialization. A layer can contain a variable number of capsules. The capsules of two consecutive layers are interdependent, which implies the fact that the activations of a layer are based on the output of the previous layer.

\subsubsection{Hierarchical Attention Network (HAN)}
HANs \cite{yang:2016} are designed to capture details about phrase structures. They construct a phrase representation by aggregating sentence representations and, for the next level, word representations. An attention mechanism is focused on identifying the key elements in the phrase which can help in boosting the overall performance. This is why the authors use two levels of attention, one for words and one for sentences. The HAN is built by stacking a group of layers. Firstly, a GRU-based sequence encoder is used for tracking the state of the sequences without using different memory cells. Then, the hierarchical attention comes into play, with attention mechanisms for words and sentences, respectively. After that, an output layer for the class prediction is added. We also use a BiLSTM at the character level, such that a word will be traversed twice, in both ways.

In our case, the model architecture consists of an \textit{input layer} which takes a 1-D input of a preset dimension, the maximum length of a text, representing one entry from the training set.
The input layer is followed by an \textit{embedding layer}, in which we load the embedding matrix obtained from the mapping process we previously described. The input is represented by a group of n words,  
each one represented by a vector of k dimensions. This ensemble forms a k x n matrix, where n is the maximum number of allowed words.

\subsubsection{Multilingual BERT (mBERT)}
Based on the original breakthrough Transformer model, mBERT \cite{pires:2019} intends to build over its counterpart by allowing it to get good performances not only for the English language, but for over 100 other languages as well. mBERT proves to be a reliable model for multilingual tasks, including Language Identification, Language Similarity, or Word Alignment.
For our implementation, we used the base variant offered by the Huggingface library\footnote{\url{https://huggingface.co/transformers/}}. We fine-tuned the model on the training dataset and used it for sequence classification.

\subsubsection{XLM-RoBERTa}
XLM-RoBERTa~\cite{goyal:2020} is another Transformer-based multilingual model, trained with the Masked Language Model objective. The training process consists of selecting text samples from different languages and masking some tokens, such that the model will need to predict those  missing tokens.
The initial model was XLM-100. Alongside XLM-RoBERTa, both models are capable of processing 100 different languages. However, XLM-RoBERTa was trained on even more data, making it more reliable and accurate. In order to conclude that XLM-RoBERTa performs better, the authors used the GLUE benchmark~\cite{wang2018glue} and also compared the results to the ones obtained by other state-of-the-art models, on tasks like Named Entity Recognition, Cross-lingual Question Answering, or Cross-lingual Natural Language Inference. 

For our task, we fine-tuned the XLM-RoBERTa base variant on the provided dataset and performed sequence classification by using the fine-tuned model. We also used the implementation provided by the Huggingface library.

\section{Results}
\subsection{Data and Preprocessing}
The competition datasets are provided in the CONLL format. The first one, Hindi-English, contains 17,000 entries (14,000 for training and 3,000 for validation) annotated with the sentiment, either positive, negative, or neutral.
The Hindi-English dataset is balanced regarding the distribution of the sentiment labels. It contains 4,992 negative examples, 5,616 positive examples, and 6,392 neutral examples. In terms of word distribution, we can conclude that the Hindi language is dominant, with 206,897 words (36,862 unique), followed by English, with 147,027 words (33,450 unique) and universal words and symbols, with 89,695 words (2,991 unique).

Similarly, the Spanish-English dataset has a comparable structure, with 12,002 entries for training and 2,998 for validation. The classes are also balanced while the Spanish language is dominant inside the tweets.

As in online environments users tend to write in informal ways, we needed a way to clean the texts such that they can be used as an input for a neural network. 
Firstly, we transformed the text to all lowercase characters, such that two identical words with at least one character in different cases will not be treated the same. Then, we removed tags (structure like "@username") as they are not helpful for identifying the overall sentiment of the tweet. Apart from tags, we also removed the stopwords for all languages.

Furthermore, emojis represent groups of characters which do not convey useful information. However, if transformed in the right format, they can be further used for sentiment analysis. Therefore, we used a dictionary\footnote{\url{https://emojipedia.org/}} that maps each emoji to a word or group of words describing the message of the emojis.

Another issue in text preprocessing is represented by the usage of repeated characters. They are used to emphasize the user's idea or excitement but they are be perceived by our network as individual and separate words. Therefore, we decided to identify groups of three or more characters and replace them with a group of two of the same character.

Punctuation is another issue because characters such as comma or single period do not contribute to the meaning of the entry. However, question marks, exclamation marks, and ellipses can be used for detecting the sentiment, being a way to accentuate the writer's sentiment. We decided to keep them in our analysis.

\subsection{Experimental Settings}
We fine-tuned all the neural networks used in our experiments to maximize their performance. The hyper-parameters for each network architecture are available in Table \ref{table:table1}.

\begin{table}[h]

\begin{center}
\caption{\label{table:table1} Neural model hyper-parameters.}
\begin{tabular}{|l|l|}
\hline \bf Model & \bf Hyper-parameters  \\ \hline
\textbf{BiLSTM} & SpatialDropout=0.35, Layer=1, MaxPooling (default)  \\
\textbf{BiLSTM+Capsule} & SpatialDropout=0.2, NumCapsules=10, DimCapsule=16, Dropout=0.12, LR=0.001 \\
\textbf{BiGRU+Capsule} & SpatialDropout=0.2, NumCapsules=10, DimCapsule=16, Dropout=0.12, LR=0.001  \\
\textbf{HAN} & SpatialDropout=0.35, Layer=1, MaxPooling (default)  \\
\textbf{mBERT} & Optimizer=AdamW, LR=2e-5, eps=1e-8 \\
\textbf{XLM-RoBERTa} & Optimizer=AdamW, LR=2e-5, eps=1e-8  \\
\hline
\end{tabular}
\end{center}

\end{table}

In order to prevent overfitting, we concluded that a Dropout rate of 0.15 represents the best choice for our models. On the other hand, regarding the capsule implementations, 10 capsules each with a dimension of 16 neurons offered the best performance.
By applying an early stopping policy, we ran the experiments for 4 epochs in the case of the BiLSTM, BiLSTM+Capsule, BiGRU+Capsule, mBERT, and XLM-RoBERTa models, and 12 epochs for the HAN architecture.

\subsection{Experiments}
We experimented with six different neural architectures and one set of word embeddings (i.e. MUSE). Firstly, we trained the networks on the provided dataset, without any augmentation. Secondly, we augmented the training dataset by adding 4,981 additional code-mixed tweets, similarly annotated, taken from an online source\footnote{\url{https://github.com/mankadronit/Code-Mixed-Dataset/blob/master/codemixed.txt}}. The resulting dataset is then preprocessed and fed to the neural networks. 

 Tables \ref{table:table2} and
 \ref{table:table3} contain the results.
  For the Hindi-English task (see Table \ref{table:table2}), the best results are obtained by using the Transformer-based solutions. Moreover, mBERT has increased performance, surpassing the next result by a margin of 0.0103 accuracy and average F1-score of 0.0119 as well as 0.0765 accuracy and 0.0771 average F1-score compared to the BiLSTM model.
 As baselines, we used two well-known classifiers, Support Vector Machines and XGBoost~\cite{Chen_2016}, and also a CNN.
The HAN model obtains an average F1-score 0.0989 lower when compared to the mBERT architecture, but yields better results than the BiLSTM+Capsule model.

For most models, the augmented dataset helps to obtain slightly better results. Adding extra training tweets can help the network with the entire process of identifying the meaningful aspects of the phrase.

For the Spanish-English task, we only experimented with the top two models from Hindi-English (i.e, mBERT and XLM-RoBERTa) and their performance has swapped. That is, XLM-RoBERTa proves to be the most efficient, marginally surpassing mBERT by a difference of 0.002 in terms of precision, 0.0142 for recall, and 0.0105 for average F1-score (see Table \ref{table:table3}).

\begin{table}[h!]
\centering
\caption{Experimental results on the Hindi-English test set. }
\label{table:table2}
\smallskip
\begin{tabularx}{\textwidth}{ |c| *{6}{Y|} }
\hline
 \multirow{ 2}{*}{\bf Method} & \multicolumn{2}{c|}{\bf Competition Dataset}  & \multicolumn{2}{c|}{\bf \makecell{Competition Dataset +\\ Augmentation Dataset}}\\ 
\cline{2-5}
 & \bf Accuracy & \bf Average F1 & \bf Accuracy & \bf Average F1\\
\hline
 XGBOOST & 0.4399 &  0.4375 &  0.4391 & 0.4255 \\
 SVM & 0.3880 &  0.2122 &  0.3712 & 0.2078 \\
 CNN    & 0.5477 & 0.4713 & 0.5199 & 0.0940 \\
 \hline
 \hline
 BiLSTM  &  0.6080 & 0.6115 &  0.6101 &  0.6144 \\
 BiLSTM+Capsule &  0.5822 & 0.6147 & 0.5839 &  0.6063 \\
 BiGRU+Capsule &  0.5954 & 0.5860 & 0.6002 &  0.5901 \\
 HAN  &  0.5853 & 0.5855 & 0.5903 &  0.5926 \\
 mBERT & \textbf{0.6856} & \textbf{0.6901} & \textbf{0.6866}  & \textbf{0.6915} \\
 XLM-RoBERTa & 0.6716 & 0.6743 & 0.6763 & 0.6796 \\
\hline
\end{tabularx}
\end{table}

\begin{table}[h!]
\centering
\caption{Experimental results on the Spanish-English test set. }
\label{table:table3}
\smallskip
\begin{tabularx}{\textwidth}{ |c| *{3}{Y|} }

\hline
\bf Method & \bf Precision & \bf Recall & \bf Average F1 \\
\hline

 mBERT & 0.8044 & 0.6297 & 0.6959 \\
 XLM-RoBERTa & \textbf{0.8064} & \textbf{0.6439} & \textbf{0.7064}  \\
\hline
\end{tabularx}
\end{table}

Furthermore, we also experimented with another level of augmentation, a synthetic one. Thus, we identified short tweets from the dataset, with a length below a threshold (i.e., the number of words minus a quarter of its average) and combined them in pairs such that we obtained new entries. However, we needed to combine tweets that were part of the same class. For example, two negative tweets can form another negative tweet. We managed to extend the dataset to 21,919 entries (including the initial augmentation), instead of 14,000.

On the other hand, we also proposed a MTL approach by training the Transformer-based networks on a secondary task. That way, the model was able to learn additional features, in order to enhance its performance. The auxiliary task consisted of predicting the tweet's dominant language. Because the dataset came in the CONLL format, we were able to determine the main language (English or Hindi) and label the tweet accordingly. Table \ref{table:table4} shows the results obtained for this second experimental stage. Once again, the Transformer-based solution proves to offer enhanced performance, surpassing the version that does not use MTL. The average F1-score of 0.6968 obtained by the XLM-RoBERTa experiment surpasses the previous iteration by a margin of 0.0053, when compared to mBERT and 0.0172 when compared directly to XLM-RoBERTa.

\begin{table}[h!]
\centering
\caption{Experimental results on the Hindi-English test set using synthetic data and MTL (after task deadline).}
\label{table:table4}
\smallskip
\begin{tabularx}{\textwidth}{ |c| *{6}{Y|} }
\hline
\bf Method & \bf Accuracy & \bf Average F1 \\
\hline
 mBERT, synthetic  data extension & 0.6810 & 0.6857 \\
 XLM-RoBERTa, synthetic  data extension & 0.6883 & 0.6912 \\
 mBERT, MTL & 0.6755 & 0.6790 \\
 XLM-RoBERTa, MTL & \textbf{0.6905} & \textbf{0.6968} \\
\hline
\end{tabularx}
\end{table}

\section{Conclusions}
This paper presents our solutions for the code-mixed sentiment analysis shared task, organized by SemEval 2020. We experimented with state-of-the-art natural language processing models, alongside training dataset extensions and a MTL technique.
Sentiment analysis proved to become challenging if the subject data is written in a code-mixed format. Because of the difficulty imposed by detecting ideas and sentiments from texts based on two different languages, merged into a single one, it becomes a requirement to develop ways to analyze the sources.

However, the performance of our models is vastly influenced by a series of prerequisites, one of them being represented by the word embeddings. Proper word embeddings can boost performances by a large margin, considering the fact that it already offers the model an insight into that language. The problem becomes more complicated if we deal with two languages instead of one. This situation requires a modality of mapping two sets of embeddings into a single vectorial space for the solutions that are not based on Transformers.

Furthermore, various network architectures can lead to different results. Networks that intend to capture the sequentiality of the text prove to be extremely efficient, obtaining better results. At the same time, networks with layers that increase in specificity can perform certain analysis tasks for distinct levels of the text.
Additionally, XLM-RoBERTa performs better when exposed to synthetic data alongside a MTL training technique. This aspect can be attributed to its ability to better identify features inside the texts, including synthetic ones, considering that it was trained on the largest amount of data out of the models used for experiments. 

For future work, we intend to experiment with the large versions of the previously mentioned language models. Theoretically, a greater number of parameters should help us improve our performance. Moreover, we also intend to include a different auxiliary task for the MTL approach, by avoiding the hard assignment of the language class and replacing it with a soft assignment.

\bibliographystyle{coling}
\bibliography{semeval2020}

\begin{thebibliography}{}

\bibitem[\protect\citename{Bojanowski \bgroup et al.\egroup
  }2017]{bojanowski2016enriching}
Piotr Bojanowski, Edouard Grave, Armand Joulin, and Tomas Mikolov.
\newblock 2017.
\newblock Enriching word vectors with subword information.
\newblock {\em Transactions of the Association for Computational Linguistics},
  5:135--146.

\bibitem[\protect\citename{Chen and Guestrin}2016]{Chen_2016}
Tianqi Chen and Carlos Guestrin.
\newblock 2016.
\newblock Xgboost: A scalable tree boosting system.
\newblock In {\em Proceedings of the 22nd acm sigkdd international conference
  on knowledge discovery and data mining}, pages 785--794.

\bibitem[\protect\citename{Cho \bgroup et al.\egroup }2014]{cho:2014}
Kyunghyun Cho, Bart Van~Merri{\"e}nboer, Caglar Gulcehre, Dzmitry Bahdanau,
  Fethi Bougares, Holger Schwenk, and Yoshua Bengio.
\newblock 2014.
\newblock Learning phrase representations using rnn encoder-decoder for
  statistical machine translation.
\newblock {\em arXiv preprint arXiv:1406.1078}.

\bibitem[\protect\citename{Conneau \bgroup et al.\egroup
  }2017]{conneau2017word}
Alexis Conneau, Guillaume Lample, Marc'Aurelio Ranzato, Ludovic Denoyer, and
  Herv{\'e} J{\'e}gou.
\newblock 2017.
\newblock Word translation without parallel data.
\newblock {\em arXiv preprint arXiv:1710.04087}.

\bibitem[\protect\citename{Conneau \bgroup et al.\egroup }2019]{goyal:2020}
Alexis Conneau, Kartikay Khandelwal, Naman Goyal, Vishrav Chaudhary, Guillaume
  Wenzek, Francisco Guzm{\'a}n, Edouard Grave, Myle Ott, Luke Zettlemoyer, and
  Veselin Stoyanov.
\newblock 2019.
\newblock Unsupervised cross-lingual representation learning at scale.
\newblock {\em arXiv preprint arXiv:1911.02116}.

\bibitem[\protect\citename{Fukushima and Miyake}1982]{fukushima:neocognitronbc}
Kunihiko Fukushima and Sei Miyake.
\newblock 1982.
\newblock Neocognitron: A self-organizing neural network model for a mechanism
  of visual pattern recognition.
\newblock In {\em Competition and cooperation in neural nets}, pages 267--285.
  Springer.

\bibitem[\protect\citename{Hinton \bgroup et al.\egroup }2018]{e2018matrix}
Geoffrey~E Hinton, Sara Sabour, and Nicholas Frosst.
\newblock 2018.
\newblock Matrix capsules with {EM} routing.
\newblock In {\em International Conference on Learning Representations}.

\bibitem[\protect\citename{Hochreiter and Schmidhuber}1997]{hochreiter:1997}
Sepp Hochreiter and J{\"u}rgen Schmidhuber.
\newblock 1997.
\newblock Long short-term memory.
\newblock {\em Neural computation}, 9(8):1735--1780.

\bibitem[\protect\citename{Mandal and Singh}2018]{mandal:2018}
Soumil Mandal and Anil~Kumar Singh.
\newblock 2018.
\newblock Language identification in code-mixed data using multichannel neural
  networks and context capture.
\newblock In {\em Proceedings of the 2018 EMNLP Workshop W-NUT: The 4th
  Workshop on Noisy User-generated Text}, pages 116--120.

\bibitem[\protect\citename{Pascanu \bgroup et al.\egroup
  }2013]{pascanu2012difficulty}
Razvan Pascanu, Tomas Mikolov, and Yoshua Bengio.
\newblock 2013.
\newblock On the difficulty of training recurrent neural networks.
\newblock In {\em International conference on machine learning}, pages
  1310--1318.

\bibitem[\protect\citename{Patwa \bgroup et al.\egroup
  }2020]{patwa2020sentimix}
Parth Patwa, Gustavo Aguilar, Sudipta Kar, Suraj Pandey, Srinivas PYKL,
  Bj{\"o}rn Gamb{\"a}ck, Tanmoy Chakraborty, Thamar Solorio, and Amitava Das.
\newblock 2020.
\newblock Semeval-2020 task 9: Overview of sentiment analysis of code-mixed
  tweets.
\newblock In {\em Proceedings of the 14th International Workshop on Semantic
  Evaluation ({S}em{E}val-2020)}, Barcelona, Spain, December. Association for
  Computational Linguistics.

\bibitem[\protect\citename{Pires \bgroup et al.\egroup }2019]{pires:2019}
Telmo Pires, Eva Schlinger, and Dan Garrette.
\newblock 2019.
\newblock How multilingual is multilingual bert?
\newblock In {\em Proceedings of the 57th Annual Meeting of the Association for
  Computational Linguistics}, pages 4996--5001.

\bibitem[\protect\citename{Rom{\'a}n \bgroup et al.\egroup
  }2014]{Roman2014TASS2}
Julio~Villena Rom{\'a}n, Janine~Garc{\'\i}a Morera, C{\'e}sar
  de~Pablo~S{\'a}nchez, Miguel {\'A}ngel~Garc{\'\i}a Cumbreras,
  Eugenio~Mart{\'\i}nez C{\'a}mara, Alfonso~Ure{\~n}a L{\'o}pez, and Mar{\'\i}a
  Teresa~Mart{\'\i}n Valdivia.
\newblock 2014.
\newblock Tass 2014-workshop on sentiment analysis at sepln-overview.
\newblock In {\em TASS 2014-Workshop on Sentiment Analysis at SEPLN: Workshop
  proceedings: XXX Congreso de la Sociedad Espa{\~n}ola de Procesamiento de
  Lenguaje Natural SEPLN 2014}, page~1. Sociedad Espa{\~n}ola para el
  Procesamiento del Lenguaje Natural.

\bibitem[\protect\citename{Ruder}2017]{ruder2017overview}
Sebastian Ruder.
\newblock 2017.
\newblock An overview of multi-task learning in deep neural networks.
\newblock {\em arXiv preprint arXiv:1706.05098}.

\bibitem[\protect\citename{Rumelhart \bgroup et al.\egroup
  }1985]{Rumelhart1986LearningIR}
David~E Rumelhart, Geoffrey~E Hinton, and Ronald~J Williams.
\newblock 1985.
\newblock Learning internal representations by error propagation.
\newblock Technical report, California Univ San Diego La Jolla Inst for
  Cognitive Science.

\bibitem[\protect\citename{Sane \bgroup et al.\egroup }2019]{sushmita:2019}
Sushmitha~Reddy Sane, Suraj Tripathi, Koushik~Reddy Sane, and Radhika Mamidi.
\newblock 2019.
\newblock Stance detection in code-mixed hindi-english social media data using
  multi-task learning.
\newblock In {\em Proceedings of the Tenth Workshop on Computational Approaches
  to Subjectivity, Sentiment and Social Media Analysis}, pages 1--5.

\bibitem[\protect\citename{Swami \bgroup et al.\egroup
  }2018]{swami2018englishhindi}
Sahil Swami, Ankush Khandelwal, Vinay Singh, Syed~Sarfaraz Akhtar, and Manish
  Shrivastava.
\newblock 2018.
\newblock An english-hindi code-mixed corpus: Stance annotation and baseline
  system.
\newblock {\em arXiv preprint arXiv:1805.11868}.

\bibitem[\protect\citename{Szegedy \bgroup et al.\egroup
  }2016]{szegedy2015rethinking}
Christian Szegedy, Vincent Vanhoucke, Sergey Ioffe, Jon Shlens, and Zbigniew
  Wojna.
\newblock 2016.
\newblock Rethinking the inception architecture for computer vision.
\newblock In {\em Proceedings of the IEEE conference on computer vision and
  pattern recognition}, pages 2818--2826.

\bibitem[\protect\citename{Trivedi \bgroup et al.\egroup }2018]{trivedi:2018}
Shashwat Trivedi, Harsh Rangwani, and Anil~Kumar Singh.
\newblock 2018.
\newblock Iit (bhu) submission for the acl shared task on named entity
  recognition on code-switched data.
\newblock In {\em Proceedings of the Third Workshop on Computational Approaches
  to Linguistic Code-Switching}, pages 148--153.

\bibitem[\protect\citename{Vaswani \bgroup et al.\egroup
  }2017]{vaswani2017attention}
Ashish Vaswani, Noam Shazeer, Niki Parmar, Jakob Uszkoreit, Llion Jones,
  Aidan~N Gomez, {\L}ukasz Kaiser, and Illia Polosukhin.
\newblock 2017.
\newblock Attention is all you need.
\newblock In {\em Advances in neural information processing systems}, pages
  5998--6008.

\bibitem[\protect\citename{Velichkov \bgroup et al.\egroup
  }2014]{rosenthal2019semeval2014}
Boris Velichkov, Borislav Kapukaranov, Ivan Grozev, Jeni Karanesheva, Todor
  Mihaylov, Yasen Kiprov, Preslav Nakov, Ivan Koychev, and Georgi Georgiev.
\newblock 2014.
\newblock Su-fmi: System description for semeval-2014 task 9 on sentiment
  analysis in twitter.
\newblock In {\em SemEval@ COLING}, pages 590--595. Citeseer.

\bibitem[\protect\citename{Vilares \bgroup et al.\egroup }2015]{vilares:2015}
David Vilares, Miguel~A Alonso, and Carlos G{\'o}mez-Rodr{\'\i}guez.
\newblock 2015.
\newblock Sentiment analysis on monolingual, multilingual and code-switching
  twitter corpora.
\newblock In {\em Proceedings of the 6th Workshop on Computational Approaches
  to Subjectivity, Sentiment and Social Media Analysis}, pages 2--8.

\bibitem[\protect\citename{Wang \bgroup et al.\egroup }2016]{wang:2016}
Zhongqing Wang, Yue Zhang, Sophia Lee, Shoushan Li, and Guodong Zhou.
\newblock 2016.
\newblock A bilingual attention network for code-switched emotion prediction.
\newblock In {\em Proceedings of COLING 2016, the 26th International Conference
  on Computational Linguistics: Technical Papers}, pages 1624--1634.

\bibitem[\protect\citename{Wang \bgroup et al.\egroup }2018]{wang2018glue}
Alex Wang, Amanpreet Singh, Julian Michael, Felix Hill, Omer Levy, and Samuel
  Bowman.
\newblock 2018.
\newblock Glue: A multi-task benchmark and analysis platform for natural
  language understanding.
\newblock In {\em Proceedings of the 2018 EMNLP Workshop BlackboxNLP: Analyzing
  and Interpreting Neural Networks for NLP}, pages 353--355.

\bibitem[\protect\citename{Yang \bgroup et al.\egroup }2016]{yang:2016}
Zichao Yang, Diyi Yang, Chris Dyer, Xiaodong He, Alex Smola, and Eduard Hovy.
\newblock 2016.
\newblock Hierarchical attention networks for document classification.
\newblock In {\em Proceedings of the 2016 conference of the North American
  chapter of the association for computational linguistics: human language
  technologies}, pages 1480--1489.

\end{thebibliography}

\end{document}